\documentclass[conference]{IEEEtran}
\IEEEoverridecommandlockouts
\usepackage{cite}
\usepackage{amsmath,amssymb,amsfonts}
\usepackage{algorithmic}
\usepackage{graphicx}
\usepackage{array} 
\usepackage{amsmath, amssymb}
\usepackage{textcomp}
\usepackage{hyperref}
\usepackage{xcolor}
\usepackage{url}
\usepackage[caption=false,font=footnotesize]{subfig}
\def\BibTeX{{\rm B\kern-.05em{\sc i\kern-.025em b}\kern-.08em
    T\kern-.1667em\lower.7ex\hbox{E}\kern-.125emX}}

\newcommand{\NameSize}{\large}         
\newcommand{\AffilSize}{\normalsize}   
\newcommand{\EmailSize}{\small}        
\newcommand{\diag}[1]{\mathbf{diag}(#1)}
\newcommand{\NameAffilSkip}{-0.25ex}    
\newcommand{\RowGap}{0.9em}             
\begin{document}

\title{OTESGN: Optimal Transport-Enhanced Syntactic-Semantic Graph Networks for Aspect-Based Sentiment Analysis\\

\thanks{This work was supported by the National Social Science Fund of China (No. 22BTQ045).}
}

\newcommand{\GDUFSaffil}{%
  {\AffilSize
  \textit{School of Information Science and Technology}\\
  \textit{Guangdong University of Foreign Studies}\\
  Guangzhou, China}%
}
\newcommand{\GDUFSuni}{%
  {\AffilSize
  \textit{Guangdong University of Foreign Studies}\\
  Guangzhou, China}%
}

\author{%
  \begin{minipage}[t]{0.31\textwidth}\centering
    {\NameSize Xinfeng Liao}\\[\NameAffilSkip]
    \GDUFSaffil\\ {\EmailSize liaoxinfeng218@163.com}
  \end{minipage}\hfill
  \begin{minipage}[t]{0.31\textwidth}\centering
    {\NameSize Xuanqi Chen}\\[\NameAffilSkip]
    \GDUFSaffil\\ {\EmailSize chenxuanqi6@163.com}
  \end{minipage}\hfill
  \begin{minipage}[t]{0.31\textwidth}\centering
  {\NameSize Lianxi Wang\textsuperscript{*}}\\[\NameAffilSkip]
  {\AffilSize \textit{Key Laboratory of Cross-Cultural Information Analysis and Intelligent Decision-Making}}\\
  \GDUFSuni\\
  {\EmailSize wanglianxi@gdufs.edu.cn}
\end{minipage}
  \\[\RowGap]
  \begin{minipage}[t]{0.31\textwidth}\centering
    {\NameSize Jiahuan Yang}\\[\NameAffilSkip]
    \GDUFSaffil\\ {\EmailSize 2397231316@qq.com}
  \end{minipage}\hfill
  \begin{minipage}[t]{0.31\textwidth}\centering
    {\NameSize Zhuowei Chen}\\[\NameAffilSkip]
    \GDUFSaffil\\ {\EmailSize zhuowei.chen1024@outlook.com}
  \end{minipage}\hfill
  \begin{minipage}[t]{0.31\textwidth}\centering
    {\NameSize Ziying Rong}\\[\NameAffilSkip]
    \GDUFSaffil\\ {\EmailSize rongziying1016@163.com}
  \end{minipage}
  \thanks{*Corresponding author.}
}

\maketitle

\begin{abstract}
Aspect-based sentiment analysis (ABSA) aims to identify aspect terms and determine their sentiment polarity. While dependency trees combined with contextual semantics provide structural cues, existing approaches often rely on dot-product similarity and fixed graphs, which limit their ability to capture nonlinear associations and adapt to noisy contexts. To address these limitations, we propose the Optimal Transport-Enhanced Syntactic-Semantic Graph Network (OTESGN), a model that jointly integrates structural and distributional signals. Specifically, a Syntactic Graph-Aware Attention module models global dependencies with syntax-guided masking, while a Semantic Optimal Transport Attention module formulates aspect-opinion association as a distribution matching problem solved via the Sinkhorn algorithm. An Adaptive Attention Fusion mechanism balances heterogeneous features, and contrastive regularization enhances robustness. Extensive experiments on three benchmark datasets (Rest14, Laptop14, and Twitter) demonstrate that OTESGN delivers state-of-the-art performance. Notably, it surpasses competitive baselines by up to +1.30 Macro-F1 on Laptop14 and +1.01 on Twitter. Ablation studies and visualization analyses further highlight OTESGN's ability to capture fine-grained sentiment associations and suppress noise from irrelevant context.

\end{abstract}


\section{Introduction}
\begin{figure}[t]
\centering
\includegraphics[width=\linewidth]{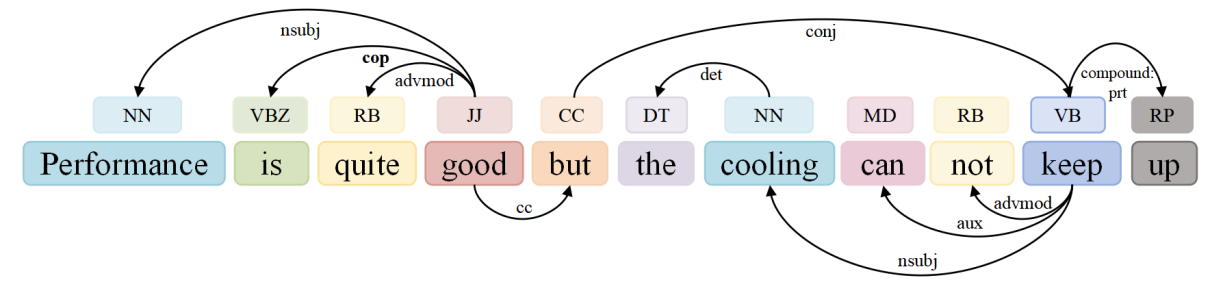}
\caption{Dependency tree example constructed with Stanford CoreNLP, featuring dual aspects (\texttt{"performance"} and \texttt{"cooling"}) exhibiting divergent sentiment polarities}
\label{fig1}
\end{figure}

Aspect-based Sentiment Analysis (ABSA) aims to identify specific aspect terms within a text and determine their corresponding sentiment polarity (e.g., positive, negative, or neutral). A key challenge lies in effectively modeling the semantic dependencies between aspect terms and their relevant contextual cues. For example, as illustrated in Fig.~\ref{fig1}, ABSA needs to recognize the aspect terms  \texttt{"performance"} and \texttt{"cooling"}, and precisely associate \texttt{"performance"} with \texttt{"quite good"} and \texttt{"cooling"} with \texttt{"not keep up"}—determining a positive sentiment for \texttt{"performance"} and an indirectly expressed negative sentiment for \texttt{"cooling"}.

In recent years, various deep neural networks and their variants have been widely applied to ABSA tasks, among which attention-based methods have become the mainstream. Traditional attention-based approaches \cite{b1,b2,b3} primarily enhance feature representation by computing semantic associations between aspect terms and opinion words, enabling the model to learn appropriate weight distributions. Prior research has demonstrated the efficacy of attention mechanisms for identifying aspect-specific importance \cite{b4}. However, these approaches frequently exhibit susceptibility to semantic interference from sentiment-irrelevant terms and often fail to adequately account for critical syntactic dependency structures. As a result, their performance is often insufficient when dealing with subtle sentiment expressions in complex contexts.

To enhance the syntactic modeling capability of attention mechanisms, researchers have proposed various solutions. For instance,  the authors of \cite{b5} introduced a multi-channel syntax-enhanced attention mechanism by incorporating syntactic relative distance (SRD) derived from the shortest dependency path to suppress syntactically irrelevant noise. On the other hand, the paper \cite{b6} modeled syntactic and semantic dependencies between words using dependency graphs. The introduction of Graph Neural Networks (GNNs) represents another major line of research. The pioneering work in \cite{b7} applied Graph Convolutional Networks (GCNs) and Graph Attention Networks (GATs) \cite{b8} to syntactic information integration, stimulating numerous subsequent studies based on GCNs \cite{b9} and GATs \cite{b10}.
Notably, the paper \cite{b11} proposes a dual-channel Transformer-enhanced architecture with gated attention, while the authors of \cite{b12}  constructed aspect-opinion dependency graphs using biaffine attention. The underlying principle unifying these approaches lies in applying differentiated weighting to dependency neighbor nodes within the graph, effectively suppressing noise and improving representation quality.

Current GCN-based methods face two key limitations. First, linear attention mechanisms based on dot-product similarity struggle to capture nonlinear semantic relationships in complex contexts, allowing critical opinion words to be overshadowed by noise from irrelevant terms \cite{b9}. Second, graph-based methods lack dynamic evolution mechanisms, as their fixed topological connections fail to adapt to input data or task demands, undermining semantic modeling accuracy \cite{b14}. Within this context, Optimal Transport (OT) offers a novel perspective by minimizing transport costs between source and target distributions under a predefined cost function. OT provides geometrically sensitive distance metrics, capturing the shape and spatial displacement of probability measure supports beyond local density comparisons. Using entropy regularization, such as the Sinkhorn algorithm \cite{b32}, OT enables efficient solutions and excels in tasks like cross-modal alignment \cite{b15} and biomolecule matching \cite{b16}.

To tackle the outlined challenges, we introduce the Optimal Transport-Enhanced Syntactic-Semantic Graph Network (OTESGN), a novel approach that implicitly models word associations via semantic distribution matching. OTESGN leverages BERT to produce context-aware word embeddings and employs dependency syntax to construct a mask matrix for enhanced structural support. A Syntactic-Semantic Collaborative Attention (SSCA) integrates local and global semantics through two components: Syntactic Graph-Aware Attention (SGAA), which captures global dependencies using syntax-guided self-attention, and Semantic Optimal Transport Attention (SOTA), which computes directed aspect-opinion word associations by solving an optimal transport coupling matrix with Sinkhorn iteration, using the inverse of cosine similarity as the transport cost. An Adaptive Attention Fusion (AAF) module dynamically balances these channels, while multi-layer residual connections update neighborhood features iteratively, with contrastive learning enhancing robustness.Experiments across three benchmark datasets show OTESGN significantly outperforms baselines in accuracy and F1 score, validating its effectiveness in aligning aspect and opinion words and its robustness to noise.

The main contributions of this paper are summarized as follows:

\begin{itemize}
    \item We propose a novel Optimal Transport-Enhanced Syntactic-Semantic Graph Network (OTESGN) for ABSA, which integrates Optimal Transport theory with graph neural networks to achieve robust semantic alignment and noise-resistant sentiment analysis.
    \item We design a Syntactic-Semantic Collaborative Attention (SSCA) that combines Syntactic Graph-Aware Attention (SGAA) and Semantic Optimal Transport Attention (SOTA), with an Adaptive Attention Fusion (AAF) module to dynamically integrate features and Contrastive Regularization to enhance robustness.
    \item We demonstrate through experiments on three benchmark datasets that OTESGN significantly outperforms state-of-the-art baselines, validating the effectiveness of Optimal Transport in improving ABSA performance.  Code and datasets available\footnote{\url{https://github.com/Chenxuanqi666/OTESGN}}.

\end{itemize}

\section{Related work}

ABSA is a fine-grained text classification task aimed at identifying the sentiment expressed toward specific aspects within a text. Early approaches based on rules \cite{b17}, sentiment lexicons, and traditional machine learning \cite{b18} offered strong interpretability but struggled to handle complex semantic structures and implicit sentiment expressions. In recent years, attention-based and graph-based methods have become widely adopted in ABSA.

\textbf{Attention-based methods:} Attention mechanisms enhance model focus on key contextual information by assigning higher weights to words relevant to aspects. Existing approaches combine self-attention \cite{b2} or interactive attention mechanisms \cite{b19} with recurrent neural networks \cite{b1,b4} or Transformers \cite{b11} to model contextual semantics. For example, the authors of \cite{b20} proposed the Interactive Attention Network (IAN), which models aspects and context separately and captures their interactions via bidirectional attention, improving sentiment polarity detection. Similarly, the paper \cite{b21} introduced a dual-level attention mechanism at both word and clause levels to attend to aspect-related keywords and clauses.

\begin{figure*}[t]
\centering
\includegraphics[width=\textwidth]{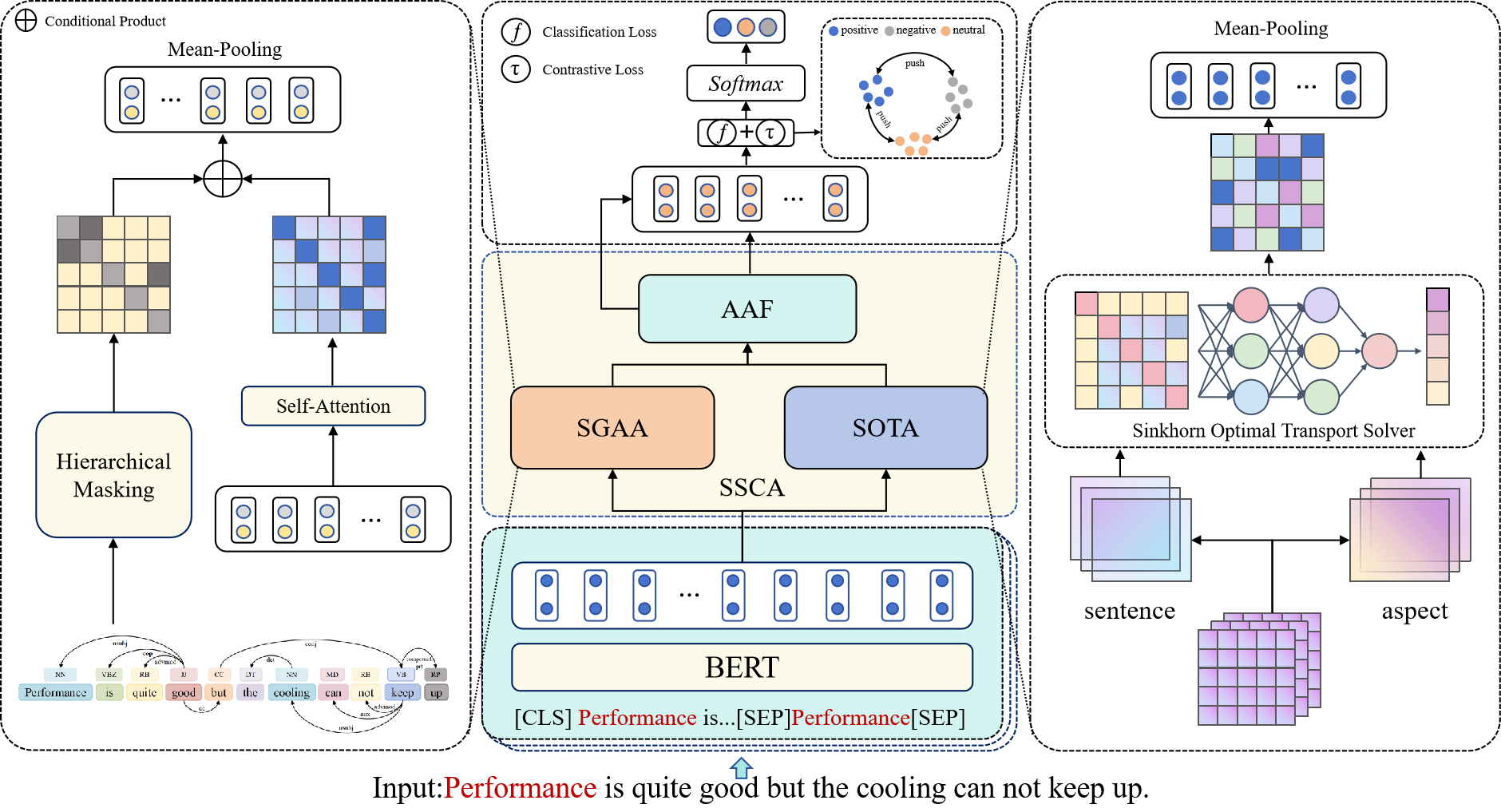}
\caption{Overall Architecture of the OTESGN Framework. Bi-Orthogonal Attention comprises two principal channels: SGAA for structural perception and SOTA for semantic transmission, with adaptive fusion mediated by the AAF}
\label{fig2}
\end{figure*}

\textbf{Graph structure-based methods:} 
Graph-based approaches have gained traction by effectively integrating syntactic dependencies and external knowledge \cite{b7}. These methods convert text into dependency or sentiment graphs and leverage Graph Convolutional Networks  \cite{b12,b14,b22,b23,b24,b25,b37} and Graph Attention Networks  \cite{b10,b26} to capture deep structural relations between aspects and context. The strong inductive bias of graph structures enhances efficient information propagation among syntax-related nodes \cite{b27,b38}. Representative models include DM-GCN \cite{b28}, which uses multi-head attention to build dependency subgraphs for enhanced semantic extraction, and APSA-GCN \cite{b29}, which employs a dual-module architecture to fuse syntactic, semantic, and lexicon features.

However, despite the strong performance of attention mechanisms on benchmark datasets, they often suffer from attention noise, which is characterized by misallocating high weights to irrelevant or neutral tokens and introducing spurious semantic associations \cite{b3}. Existing graph-based methods, while effective, typically rely on static or manually constructed graphs, which limits their adaptability across domains and hampers their ability to dynamically capture implicit semantic relations \cite{b9}. Additionally, local aggregation strategies tend to overlook global semantic alignment.

Recently, the integration of Optimal Transport with graph structures has shown great promise across various domains by jointly optimizing distribution alignment and relational modeling \cite{b30}. For instance, the authors of \cite{b15} proposed the UOT-RCL framework, which leverages OT to reduce cross-modal discrepancies via alignment. Similarly,  the paper \cite{b31} introduced the Temporal Routing Adapter (TRA), which employs OT to dynamically allocate samples to predictors under multi-pattern trading scenarios, addressing implicit routing challenges. Inspired by this, we extend OT to the ABSA domain and propose the OTESGN model, in which semantic optimal transport attention effectively aligns aspect terms with their corresponding opinion expressions.

\section{Methodology}
The proposed Optimal Transport-Enhanced Syntactic-Semantic Graph Network (OTESGN), depicted in Fig.~\ref{fig2}, is structured around a methodological framework comprising four principal stages: input encoding, syntactic-semantic collaborative attention, progressive aspect-aware learning, and multi-objective training.

\subsection{Input Encoding Layer}
Given a text-aspect pair \((s, a)\), where the text \(s = \{w_1, w_2, \dots, w_n\}\) contains \(n\) tokens and the aspect terms \(a = \{a_1, a_2, \dots, a_m\}\) form a contiguous subsequence of \(s\) (\(1 \leq m \leq n\)), we adopt a pre-trained BERT model (e.g., BERT-base-uncased) with its WordPiece tokenizer for the input format: \texttt{[CLS] + Tokenize(}$s$\texttt{) + [SEP] + Tokenize(}$a$\texttt{) + [SEP]}.

Deep contextualized representations are obtained from the BERT model. The hidden representation of the full sequence is:
\begin{equation}
\begin{aligned}
H = \mathit{BERT}_{\mathit{Encoder}} \Big( &[w_{\mathit{[CLS]}}, w_1, \dots, w_n, \\
&w_{\mathit{[SEP]}}, w_{a_1}, \dots, w_{a_m}, w_{\mathit{[SEP]}}] \Big)
\end{aligned}
\end{equation}
yielding the contextualized representation \( H = \{ h_0, h_1, \dots, h_{n+1}, h_{n+2}, \dots, h_{n+m+2} \} \in \mathbb{R}^{(n+m+3) \times d} \), 
where indices start from 0 for [CLS], 1 to \(n\) for \(s\), \(n+1\) for the first [SEP], \(n+2\) to \(n+m+1\) for \(a\), and \(n+m+2\) for the final [SEP]. For subsequent processing, we extract the text-specific hidden representations \( H^s = \{ h_1, h_2, \dots, h_n \} \in \mathbb{R}^{n \times d} \). The aspect-specific hidden subvector sequence is denoted as 
\( h_a = \{ h_{pos(a_1)}, h_{pos(a_2)}, \dots, h_{pos(a_m)} \} \in \mathbb{R}^{m \times d} \), where \( pos(a_j) \) are the indices of aspect tokens in \(s\) (i.e., 1 to \(n\)).

\subsection{Syntactic-Semantic Collaborative Attention}
Determining the sentiment polarity of an aspect term necessitates modeling both its syntactic relations and semantic dependencies with contextual tokens. Based on the assumption that sentiment relies on local and global syntactic cues, previous methods often use dependency trees to model syntax and attention mechanisms to link aspects with their context. Inspired by this, we propose a \textbf{Syntactic-Semantic Collaborative Attention (SSCA)} mechanism that integrates \textbf{Syntactic Graph-Aware Attention (SGAA)} and \textbf{Semantic Optimal Transport Attention (SOTA)} to jointly capture structural and semantic dependencies. The collaborative design leverages the conceptual orthogonality between syntactic (graph-based) and semantic (distributional) cues, implemented via weighted fusion to achieve accurate aspect-opinion alignment, thereby better identifying the true sentiment orientation of aspect terms.

\subsubsection{Syntactic Graph-Aware Attention}
Traditional methods relying solely on syntactic parsing or isolated attention mechanisms often fail to capture implicit semantic associations and are sensitive to irrelevant tokens. To address these issues, we propose a \textbf{SGAA} mechanism that synergistically integrates syntactic constraints with attentional flexibility. SGAA constructs a syntactic adjacency matrix from the dependency parse tree to restrict attention propagation between unrelated words while employing self-attention to adaptively weigh dependency strengths, capturing nuanced semantic relationships and minimizing noise from irrelevant edges.

Following SSEGCN~\cite{b9}, we model the syntactic dependency tree as an undirected graph to simplify shortest-path computation, with distances computed using Stanford CoreNLP. The distance between nodes \( w_i \) and \( w_j \), denoted \( d(w_i, w_j) \), is the number of edges in the shortest path:
\begin{equation}
D(i,j) = d(w_i, w_j).
\end{equation}

Using the shortest path distance, we construct a set of multi-granularity mask matrices \( M = \{ M^{1}, M^{2}, \ldots, M^{p} \} \), where \( p \) is the number of attention heads and each \( M^{k} \in \mathbb{R}^{n \times n} \). The mask matrix for the \( k \)-th head is defined as:
\begin{equation}
M^{k}_{ij} = 
\begin{cases}
0, & \text{if } D(i,j) \leq \tau^{k} \\
-\infty, & \text{otherwise}
\end{cases},
\end{equation}
where \( \tau^{k} \in [1, p] \) is the syntactic distance threshold for the \( k \)-th attention head, set linearly from 1 to \(p\) (e.g., \(\tau^k = k\)) to capture varying dependency ranges. A value of 0 permits attention between syntactically related words \( w_i \) and \( w_j \), while \( -\infty \) (implemented as a large negative value in practice) suppresses attention for unrelated pairs. Smaller \( \tau^{k} \) focuses on local dependencies, while larger \( \tau^{k} \) captures global dependencies, enabling coarse-to-fine structural constraints.

Subsequently, a multi-head attention mechanism captures word interactions:
\begin{equation}
Q = H^s W_Q, \quad K = H^s W_K,
\end{equation}
\begin{equation}
A_{SG}^k = \text{softmax}\left( \frac{Q K^T}{\sqrt{d}} + M^k \right),
\end{equation}
where \( k \in [1, p] \) denotes the \( k \)-th attention head, and queries \( Q \) and keys \( K \) are derived by projecting the BERT-derived text representations \( H^s \) using learnable weight matrices \( W_Q, W_K \in \mathbb{R}^{d \times d} \). Each head corresponds to a distinct syntactic granularity, combining global dependency modeling with syntactic constraints for flexible contextual analysis.

\begin{figure}[t]
\centering
\includegraphics[width=\linewidth]{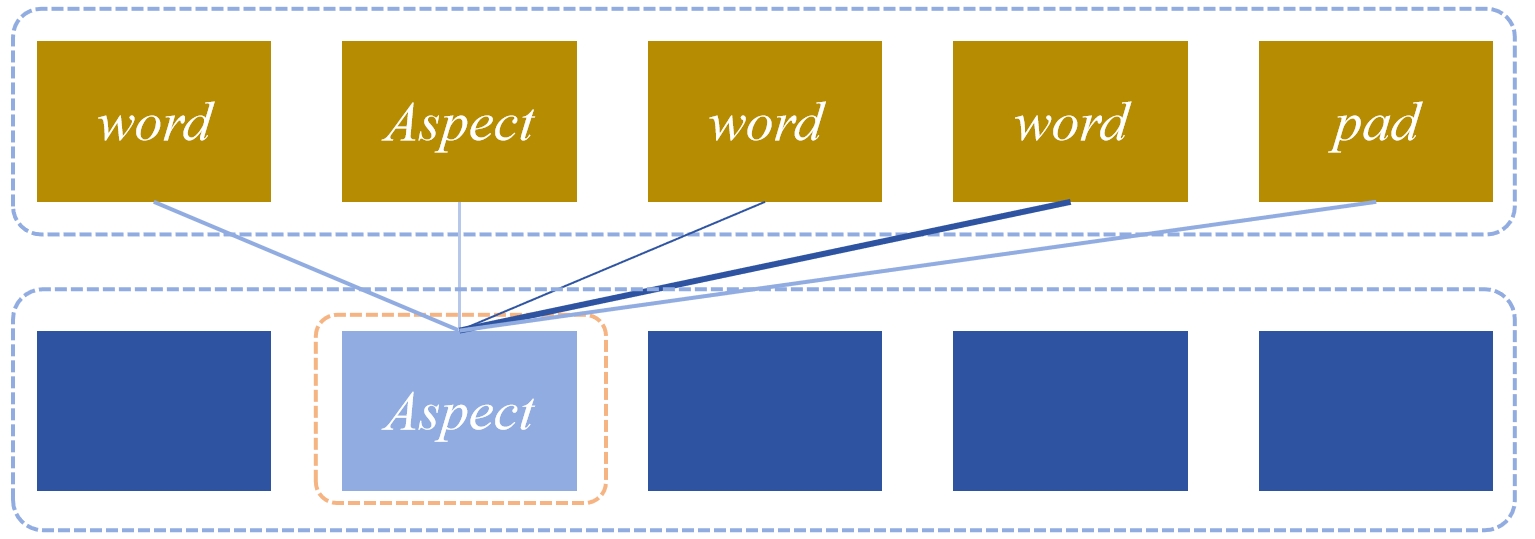}
\caption{Semantic Distance Modeling for Optimal Transport}
\label{fig3}
\end{figure}

\subsubsection{Semantic Optimal Transport Attention}
Existing methods often model semantic associations between aspect terms and their opinions using linear similarity measures, which struggle to capture complex nonlinear dependencies. To address this, we propose a \textbf{SOTA} module that frames semantic association as an optimal transport (OT) problem. In this approach, the context embeddings (text words) form the source distribution, while the aspect embedding represents the target distribution. By minimizing the semantic transport cost, the model learns the optimal alignment from context words (potential opinions) to the aspect, effectively capturing one-to-many alignment patterns in real-world texts. For multi-word aspects, average pooling reduces to many-to-one aggregation, which may lose fine-grained alignments; future work could extend to many-to-many OT by defining \(\nu \in \Delta^m\) and a cost matrix \(\text{Cost} \in \mathbb{R}^{n \times m}\).

\paragraph{Semantic Distance Modeling}
To facilitate the distributed transfer of semantic associations, it is essential to first establish the cost of transferring semantic information from the source distribution (i.e., contextual embeddings) to the target entity (i.e., aspect terms), as illustrated in Fig.~\ref{fig3}. Specifically, the hidden representations of aspect terms, derived from the output of the Transformer layer, are aggregated through average pooling to form the aspect semantic center, defined as:
\begin{equation}
h_a' = \frac{1}{m} \sum_{j=1}^{m} h_{pos(a_j)}.
\end{equation}

Subsequently, a cost matrix is constructed by computing the cosine distance between each word in the text and the aspect semantic center:
\begin{equation}
\text{Cost} = 1 - \frac{H^s (h_a')^T}{\|H^s\| \cdot \|h_a'\|},
\end{equation}
where cosine distance quantifies semantic dissimilarity, with lower costs indicating higher relevance for transport.

\paragraph{Learning Probability Distributions}
To satisfy OT constraints, we represent the source (context words) and target (aspect) as probability distributions:
\begin{equation}
\begin{aligned}
\mu &= \text{softmax}(F_{\mu}(H^s)) , \\[0.5em]
\nu &= [1] \in \Delta^{1},
\end{aligned}
\end{equation}
where \( F_{\mu} \) is a feedforward network (e.g., linear layer) projecting \( H^s \in \mathbb{R}^{n \times d} \) to \( \mathbb{R}^{n \times 1} \), followed by softmax to ensure \( \sum \mu = 1 \). The target \( \nu = [1] \) is fixed, as the aspect is a single aggregated point.

\begin{figure}[t]
\centering
\includegraphics[width=\linewidth]{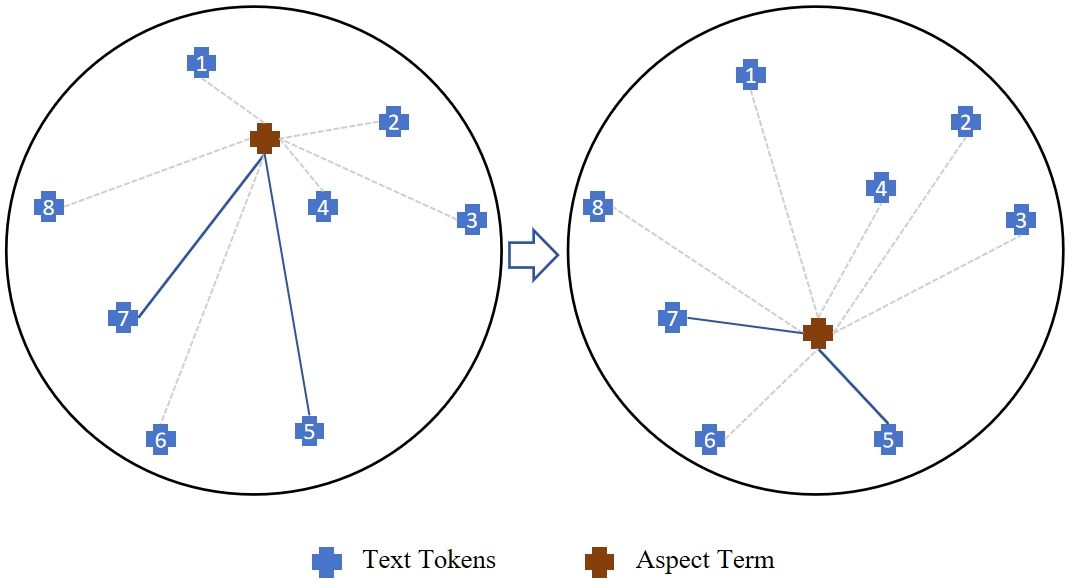}
\caption{Solving Semantic Optimal Transport in the embedding space, where the aspect term is optimally transported toward high-weight context words to achieve fine-grained semantic alignment}
\label{fig4}
\end{figure}

\paragraph{Sinkhorn Optimal Transport Solver}
To derive the optimal transport plan, which aligns the aspect term with high-weight context words in the embedding space (as depicted in Fig.~\ref{fig4}), We employ the Sinkhorn algorithm~\cite{b32} with entropic regularization. For each attention head $k \in [1, p]$, a head-specific regularization 
coefficient $\epsilon^k$ is assigned according to a predefined 
hyperparameter schedule (e.g., $\epsilon^k = 0.3 + 2.7 \cdot \frac{k-1}{p-1}$), 
thereby introducing systematic diversity in alignment granularity:
\begin{equation}
K^k = \exp\left(-\frac{\text{Cost}}{\epsilon^k}\right),
\end{equation}
where \( \epsilon^k \) controls the sharpness of the transport plan, with smaller values emphasizing precise alignments and larger values promoting smoother distributions. The dual variables \( u \in \mathbb{R}^n \) and \( v \in \mathbb{R}^1 \) are initialized as ones and iteratively updated for \( L \) iterations:
\begin{equation}
u \leftarrow \frac{\mu}{K^k v}, \quad v \leftarrow \frac{\nu}{(K^k)^T u}.
\end{equation}

The transport plan is:
\begin{equation}
\pi^k = \diag{u} K^k v,
\end{equation}
where \( \diag{u} \in \mathbb{R}^{n \times n} \) is the diagonal matrix with vector \( u \) on its diagonal. The resulting semantic attention weights are:
\begin{equation}
A_{OT}^k = \pi^k,
\end{equation}
quantifying the contribution of each context word to the aspect for fine-grained alignment.

\paragraph{Adaptive Attention Fusion}
To integrate syntactic and semantic information, we fuse the structure-aware graph attention \( A_{SG}^k \in \mathbb{R}^{n \times n} \) with the semantic optimal transport attention \( A_{OT}^k \in \mathbb{R}^{n} \). We first broadcast \( A_{OT}^k \) to a matrix form by replicating it across columns: \( A_{OT\text{-}mat}^k[i,j] = A_{OT}^k[i] \) for all \( j \), resulting in \( A_{OT\text{-}mat}^k \in \mathbb{R}^{n \times n} \). This broadcasting emphasizes rows corresponding to high-relevance context words.

The weighted fusion per head is:
\begin{equation}
A^k = \beta \cdot A_{SG}^{k} + (1 - \beta) \cdot A_{OT\text{-}mat}^{k},
\end{equation}
where \( \beta \in [0, 1] \) is a learnable scalar (initialized at 0.5) balancing the contributions. The final attention matrix averages over heads:
\begin{equation}
A = \frac{1}{p} \sum_{k=1}^{p} A^k.
\end{equation}

\subsection{Progressive Aspect-aware Learning}
The fused attention matrix \( A \in \mathbb{R}^{n \times n} \) incorporates both syntactic and semantic alignments via the specified broadcasting and fusion. To reduce computational overhead while effectively propagating multi-hop dependencies, \( A \) is static across layers. The text features from the encoder, denoted as \( H^s = \{ h_1^s, h_2^s, \ldots, h_n^s \} \), serve as the initial hidden representations, i.e., \( h^0 = H^s \). For the \( l \)-th layer, the update rule is:
\begin{equation}
h^l = h^{l-1} + \sigma(A h^{l-1} + b^l),
\end{equation}
where \( b^l \) is the bias term, and \( \sigma \) is the ReLU activation function, ensuring non-negative feature updates. The node representations from the final layer are denoted as \( h^L = \{ h^L_1, h^L_2, \ldots, h^L_n \} \). A masking mechanism retains only the representations of aspect term nodes, resulting in \( h^L_a = \{ h^L_{pos(a_1)}, h^L_{pos(a_2)}, \ldots, h^L_{pos(a_m)} \} \). Mean pooling is performed over the aspect word embeddings:
\begin{equation}
h^{pool}_a = \frac{1}{m} \sum_{j=1}^m h^L_{pos(a_j)}.
\end{equation}
Finally, \( h^{pool}_a \) is processed through a fully connected layer followed by a softmax function to produce the sentiment polarity distribution:
\begin{equation}
p(c|a) = \text{softmax}(W_p h_a^{pool} + b_p).
\end{equation}

\subsection{Multi-objective Training}
To jointly optimize sentiment classification accuracy and the discriminability of feature representations, we construct a multi-objective training framework combining the cross-entropy loss \(\mathcal{L}_e(\theta)\) with the contrastive learning loss \(\mathcal{L}_c(\theta)\):
\begin{equation}
\mathcal{L}(\theta) = \mathcal{L}_e(\theta) + \lambda \mathcal{L}_c(\theta),
\end{equation}
where \(\lambda\) (e.g., 0.1) is a hyperparameter weighting the contrastive loss. The cross-entropy loss \(\mathcal{L}_e(\theta)\) supervises sentiment polarity prediction:
\begin{equation}
\mathcal{L}_e(\theta) = - \sum_{(s,a) \in \mathcal{D}}{{\sum_{c \in C}} y_c \log p(c|a)},
\end{equation}
where \(\mathcal{D}\) denotes the set of all text–aspect term pairs, \(C\) represents the set of sentiment polarity categories, \(y_c \in \{0, 1\}\) is the ground-truth label indicator for class \(c\), and \(p(c \mid a)\) is the model-predicted probability that the aspect term \(a\) belongs to class \(c\). The contrastive learning loss \(\mathcal{L}_c(\theta)\) imposes consistency constraints on the representation space, encouraging semantically similar samples to be closer while distancing dissimilar ones:
\begin{equation}
\mathcal{L}_c(\theta) = -\frac{1}{K} \sum_{i \in \mathcal{I}} \log \frac{\exp(\text{sim}(h^{pool}_{a_i}, h^{pool}_{a_i^+})/\tau)}{\sum_{j \in \mathcal{I}} \exp(\text{sim}(h^{pool}_{a_i}, h^{pool}_{a_j})/\tau)},
\end{equation}
where \(K\) is the batch size, \(\mathcal{I}\) is the index set of all samples, and \(\text{sim}(x, y) = \frac{x^T y}{\|x\| \cdot \|y\|}\) is the cosine similarity function. Positive samples \(h^{pool}_{a_i^+}\) share the same sentiment label as \(h^{pool}_{a_i}\), while negative samples \(h^{pool}_{a_j}\) have different labels. The scalar \(\tau\) (e.g., 0.1) controls the sharpness of the similarity distribution. Hyperparameters \(\lambda\) and \(\tau\) are tuned via grid search.

\section{Experiment}

\subsection{Datasets}

This study conducts experiments on three benchmark datasets in the ABSA domain, including the restaurant dataset (Rest14) and laptop dataset (Laptop14) released by SemEval 2014 Task 4 \cite{b17}, and the Twitter dataset derived from social media posts \cite{b18}. We utilize Stanford CoreNLP to construct dependency parse trees for each dataset, which are then treated as undirected graphs. Dependency Distance (DD) is computed based on the linear positional difference between words.

\begin{figure}[t]
\centering
\includegraphics[width=\linewidth]{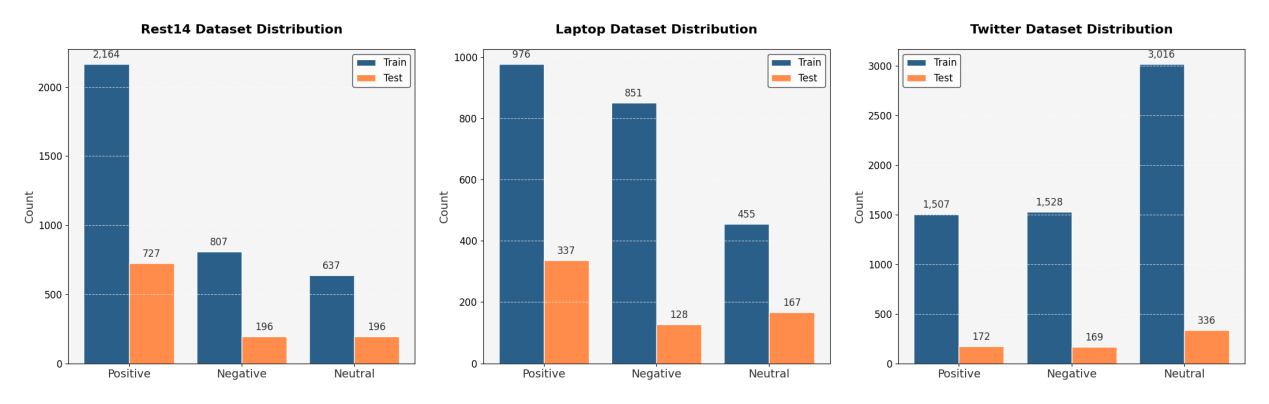}
\caption{Sample Distribution Statistics of Benchmark Datasets}
\label{fig5}
\end{figure}

The sample distribution and DD statistics across datasets are illustrated in Fig.~\ref{fig5} and Table 1. As shown in Table 1, the average DD values for Laptop14 (3.21) and Rest14 (3.17) are relatively close, while Twitter exhibits a slightly higher average of 3.41. This indicates notable linguistic variation across domains: the informal expressions and complex metaphors common in Twitter data lead to increased syntactic complexity, whereas the more formal and domain-specific terminology in Laptop14 and Rest14 contributes to more uniform syntactic structures. These findings suggest that robust models must be resilient to noisy and diverse linguistic patterns.

\subsection{Baselines}

\begin{table}[t]
\caption{Statistics of the Benchmark Datasets}
\centering
\renewcommand{\arraystretch}{1.3} 
\begin{tabular}{l|c|c|c}
\hline
\textbf{Dataset} & \textbf{Average DD ± Std} & \textbf{DD Range} & \textbf{Number of texts} \\ \hline
Laptop14 & 3.21 ± 0.85 & [1.00, 8.68] & 1863 \\ \hline
Rest14 & 3.17 ± 0.85 & [0.00, 8.43] & 2579 \\ \hline
Twitter & 3.41 ± 0.98 & [0.50, 11.97] & 6728 \\ \hline
\end{tabular}
\label{tab1}
\end{table}
To comprehensively evaluate the effectiveness of the proposed model, we select a diverse set of representative baseline models from the ABSA domain, categorized as follows:

\begin{itemize}
    \item \textbf{Traditional Neural Network Models}:
    \begin{itemize}
        \item \textit{RAM}~\cite{b1}: Leverages recurrent neural networks with a memory mechanism to model aspect-aware attention.
        \item \textit{TNet}~\cite{b2}: Integrates BiLSTM and CNN to enhance contextual representation through transformation layers.
    \end{itemize}

    \item \textbf{Pretrained Language Models}:
    \begin{itemize}
        \item \textit{BERT}~\cite{b33}: Encodes both the text and aspect term in a unified framework.
    \end{itemize}

    \item \textbf{Graph-Based Models}:
    \begin{itemize}
        \item \textit{ASGCN}~\cite{b7}, \textit{SSEGCN}~\cite{b9}, \textit{DGEDT}~\cite{b11}, \textit{CSADGCN}~\cite{b12}, \textit{DualGCN}~\cite{b13}, \textit{KDGN}~\cite{b14}, and \textit{RDGCN}~\cite{b24}: These models leverage dependency syntax or semantic graphs. Some incorporate graph attention mechanisms, orthogonality constraints, or external domain knowledge, and all utilize BERT for text encoding.
    \end{itemize}

    \item \textbf{Contrastive Learning Models}:
    \begin{itemize}
        \item \textit{SVCCL}~\cite{b25}, \textit{APSCL}~\cite{b34}, and \textit{CADA}~\cite{b35}: These models introduce supervised contrastive loss functions or data augmentation strategies to enhance the discriminability and robustness of representation learning.
    \end{itemize}

    \item \textbf{Structure-Enhanced and Hybrid Models}:
    \begin{itemize}
        \item \textit{YORO}~\cite{b22}, \textit{IDGNN}~\cite{b23}, \textit{HPEP-GCN}~\cite{b27}, and \textit{LWEDA-ACT}~\cite{b36}: These models enhance ABSA performance through various strategies, including the incorporation of structured graph relations, multi-level prompt learning, dual-channel fusion of semantic and syntactic features, and sentiment word expansion mechanisms.
    \end{itemize}
\end{itemize}

\subsection{Training Configuration}
Macro-F1 and Accuracy are adopted as evaluation metrics to comprehensively assess the model's overall performance and robustness to class imbalance in the classification task.
Word embeddings are initialized using a pre-trained BERT model. The maximum input text length is set to 100 tokens. The attention layer has a hidden dimension size of 200 and employs a multi-head attention mechanism with 5 parallel heads. The model stacks 6 of these attention layers.
All models are trained using the Adam optimizer with a batch size of 32 and a learning rate of 2e-5. To prevent overfitting, dropout regularization (rate = 0.1) is applied to the syntactic tree mask matrix and residual connections across all layers, additionally, the threshold for the syntactic matrix mask is set to 5.
For the SOTA, the entropy regularization coefficient is set to 1, and the Sinkhorn algorithm runs for a maximum of 50 iterations.


\subsection{Main Results}

\begin{table}[t]
\caption{Comparison of OTESGN with Baseline Models on Rest14, Laptop14, and Twitter Datasets. Best results are colored in \textbf{\textcolor{red}{first}}, \textbf{\textcolor{blue}{second}}, \textbf{\textcolor[rgb]{0,0.5,0}{third}} (Unit:\%)}
\begin{center}
\renewcommand{\arraystretch}{1.3} 
\begin{tabular}{>{\centering\arraybackslash}m{2.5cm}@{\hspace{0.3cm}}|c@{\hspace{0.15cm}}c@{\hspace{0.3cm}}|c@{\hspace{0.15cm}}c@{\hspace{0.3cm}}|c@{\hspace{0.15cm}}c}
\hline
\hline
\textbf{Model} & \multicolumn{2}{c|}{\textbf{Rest14}} & \multicolumn{2}{c|}{\textbf{Laptop14}} & \multicolumn{2}{c}{\textbf{Twitter}} \\
\cline{2-7}
 & \textbf{\textit{ACC}$\uparrow$} & \textbf{\textit{F1}$\uparrow$} & \textbf{\textit{ACC}$\uparrow$} & \textbf{\textit{F1}$\uparrow$} & \textbf{\textit{ACC}$\uparrow$} & \textbf{\textit{F1}$\uparrow$} \\
\hline
RAM (2017) & 80.23 & 70.80 & 74.49 & 71.35 & 69.36 & 67.30 \\
TNet (2018) & 80.69 & 71.27 & 76.54 & 71.75 & 74.90 & 73.60 \\
BERT (2019) & 85.97 & 80.09 & 79.91 & 76.00 & 75.92 & 75.18 \\
\hline
ASGCN (2019) & 80.77 & 72.02 & 75.55 & 71.05 & 72.15 & 70.40 \\
DGEDT (2020) & 86.30 & 80.00 & 79.80 & 75.60 & 77.90 & 75.40 \\
DualGCN (2021) & 87.13 & 81.16 & 81.80 & 78.10 & 77.40 & 76.02 \\
SSEGCN (2022) & 87.31 & 81.09 & 81.01 & 77.96 & 77.40 & 76.02 \\
RDGCN (2024) & \textbf{\textcolor{blue}{87.49}} & 81.16 & \textbf{\textcolor{blue}{82.12}} & 78.29 & \textbf{\textcolor{blue}{78.29}} & \textbf{\textcolor[rgb]{0,0.5,0}{77.14}} \\
CSADGCN (2024) & \textbf{\textcolor[rgb]{0,0.5,0}{87.40}} & 81.56 & \textbf{\textcolor{blue}{82.12}} & \textbf{\textcolor{blue}{79.22}} & 76.81 & 75.67 \\
\hline
APSCL (2023) & 86.86 & 81.28 & 81.02 & 78.47 & - & - \\
SVCCL (2025) & 86.34 & 80.16 & 81.35 & 77.76 & 76.72 & 75.92 \\
CADA (2025) & 87.14 & 81.26 & 80.88 & 77.71 & - & - \\
\hline
KDGN (2023) & 87.01 & \textbf{\textcolor{blue}{81.94}} & 81.32 & 77.59 & 77.64 & 75.55 \\
YORO (2024) & 87.14 & \textbf{\textcolor[rgb]{0,0.5,0}{81.83}} & 81.82 & 78.32 & - & - \\
IDGNN (2024) & 87.25 & 81.16 & 81.12 & 78.46 & 76.70 & 75.90 \\
HPEP-GCN (2025) & 86.86 & 80.65 & \textbf{\textcolor[rgb]{0,0.5,0}{81.96}} & \textbf{\textcolor[rgb]{0,0.5,0}{79.10}} & - & - \\
LWEDA-ACT (2025) & \textbf{\textcolor{red}{88.11}} & \textbf{\textcolor{red}{82.12}} & 81.17 & 78.29 & \textbf{\textcolor[rgb]{0,0.5,0}{78.17}} & \textbf{\textcolor{blue}{77.16}} \\
\hline
\textbf{OTESGN} & 87.21 & 80.47 & \textbf{\textcolor{red}{82.86}} & \textbf{\textcolor{red}{80.52}} & \textbf{\textcolor{red}{78.75}} & \textbf{\textcolor{red}{78.17}} \\
\hline
\hline
\end{tabular}
\label{tab2}
\end{center}
\end{table}

As shown in Table~\ref{tab2}, OTESGN achieves the best or competitive performance on all three benchmarks.
It attains \textbf{80.52} Macro-F1 on \textsc{Laptop14} and \textbf{78.17} on \textsc{Twitter}, improving over the strongest prior models by \textbf{+1.30 pp} and \textbf{+1.01 pp}, respectively.
On \textsc{Rest14}, OTESGN is comparable to the best baseline. Improvements are stated in percentage points (pp).

\paragraph{Versus traditional and pretrained baselines.}
Relative to RAM and TNet~\cite{b1,b2}, OTESGN yields higher Macro-F1 (80.47/80.52/78.17 on \textsc{Rest14}/\textsc{Laptop14}/\textsc{Twitter}).
Compared with BERT~\cite{b33}, it improves by \textbf{+4.52 pp}/\textbf{+3.00 pp} on \textsc{Laptop14}/\textsc{Twitter}, suggesting that incorporating explicit structural priors and optimal-transport alignment benefits aspect–opinion localization.

\paragraph{Versus graph-based models.}
On \textsc{Laptop14}, OTESGN surpasses RDGCN and KDGN~\cite{b24,b14} (80.52 vs. 78.29/77.59 Macro-F1).
We attribute the gains to hierarchical syntactic masking and gated fusion that alleviate long-range dependency decay and reduce single-tree bias observed in ASGCN~\cite{b7}.
On \textsc{Twitter}, OTESGN reaches 78.17 Macro-F1 (\textbf{+2.50 pp} over CSADGCN~\cite{b12}), indicating robustness under informal text.

\paragraph{Versus contrastive-learning models.}
While APSCL~\cite{b34} improves over DualGCN~\cite{b13} on \textsc{Laptop14} (78.47 vs. 78.10 Macro-F1), OTESGN attains \textbf{80.52} (\textbf{+2.05 pp} over APSCL).
The results support that optimal-transport–enhanced contrast (with cost-aware alignment and hard-negative mining) further refines the feature space.
On \textsc{Rest14}, OTESGN obtains 80.47 Macro-F1 (\textbf{+0.31 pp} over SVCCL~\cite{b25}).

\begin{table}[t!]
\setlength{\belowcaptionskip}{-10pt} 
\caption{Ablation Study of OTESGN on Rest14, Laptop14, and Twitter Datasets. 
“w/o SM”, “w/o GA”, “w/o OT”, “w/o GA+OT”, and “w/o CL” denote removing the Syntactic Mask, SGAA, SOTA, both SGAA \& SOTA, and Contrastive Learning, respectively (Unit:\%)}
\centering
\renewcommand{\arraystretch}{1.25}
\setlength{\tabcolsep}{4pt}
\begin{tabular}{>{\centering\arraybackslash}m{3cm}|cc|cc|cc}
\hline
\hline
\textbf{Model Variant} & \multicolumn{2}{c|}{\textbf{Rest14}} & \multicolumn{2}{c|}{\textbf{Laptop14}} & \multicolumn{2}{c}{\textbf{Twitter}} \\
\cline{2-7}
& \textit{Acc$\uparrow$} & \textit{F1$\uparrow$} & \textit{Acc$\uparrow$} & \textit{F1$\uparrow$} & \textit{Acc$\uparrow$} & \textit{F1$\uparrow$} \\
\hline
OTESGN (Full) & \textbf{\textcolor{red}{87.21}}& \textbf{\textcolor{red}{80.47}}& \textbf{\textcolor{red}{82.86}}& \textbf{\textcolor{red}{80.52}}& \textbf{\textcolor{red}{78.75}}& \textbf{\textcolor{red}{78.17}}\\
\quad w/o SM & 82.91 & 76.00 & 78.81 & 78.23 & 71.62 & 69.20 \\
\quad w/o GA & 83.94 & 77.35 & 81.66 & 79.41 & 76.37 & 74.15 \\
\quad w/o OT & 80.60 & 71.52 & 79.93 & 76.54 & 73.25 & 72.01 \\
\quad w/o GA+OT & 81.19 & 74.45 & 77.58 & 75.17 & 70.15 & 68.07 \\
\quad w/o CL & 85.74 & 79.26 & 81.27 & 78.26 & 75.19 & 73.52 \\
\hline
\hline
\end{tabular}
\label{tab3}
\end{table}

\begin{figure}[t]
\setlength{\abovecaptionskip}{0pt} 
\setlength{\belowcaptionskip}{-10pt} 
\centering
\includegraphics[width=1\columnwidth, trim=0 0 0 8pt, clip]{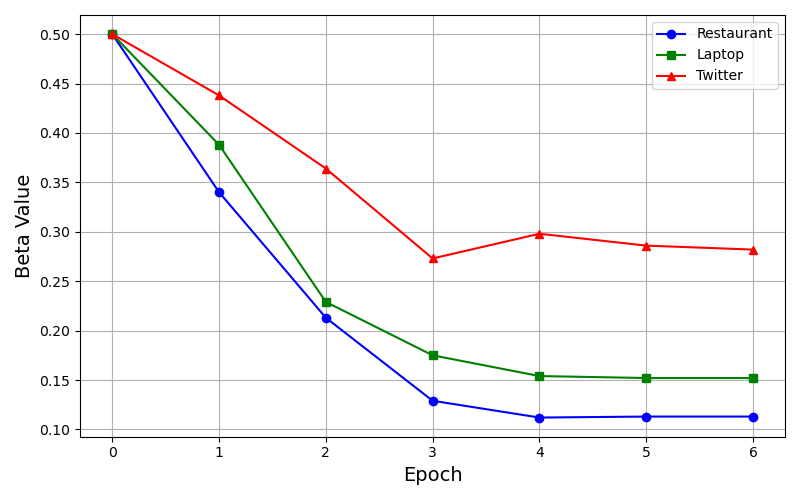} 
\caption{The evolution of the learnable parameter \(\beta\) across training epochs begins to converge to a stable value from the 4th epoch}
\label{fig6}
\end{figure}

\begin{table*}[t]
\caption{Example Sentiment Analysis}
\begin{center}
\renewcommand{\arraystretch}{1.5} 
\begin{tabular}{>{\centering\arraybackslash}m{1cm}|>{\centering\arraybackslash}m{1cm}|>{\centering\arraybackslash}p{13cm}}
\hline
\hline
\textbf{Label} & \textbf{OTESGN} & \textbf{text} \\
\hline
positive & neutral & what is the best \textbf{\textcolor{red}{harry potter}} movie for you ?\\
negative & neutral & \textbf{\textcolor{red}{nicolas cage}} is selling another castle , maybe he should be a real estate .\\
negative & positive & i wonder if people like the wiz , \textbf{\textcolor{red}{taylor swift}} , and bieber will be popular in 5 years ( i hope not )\\
\hline
\hline
\end{tabular}
\label{tab4}
\end{center}
\end{table*}

\begin{figure*}[t]
\centering
\subfloat[Laptop Confusion Matrix\label{fig:laptop_confusion}]{
    \includegraphics[width=0.3\textwidth]{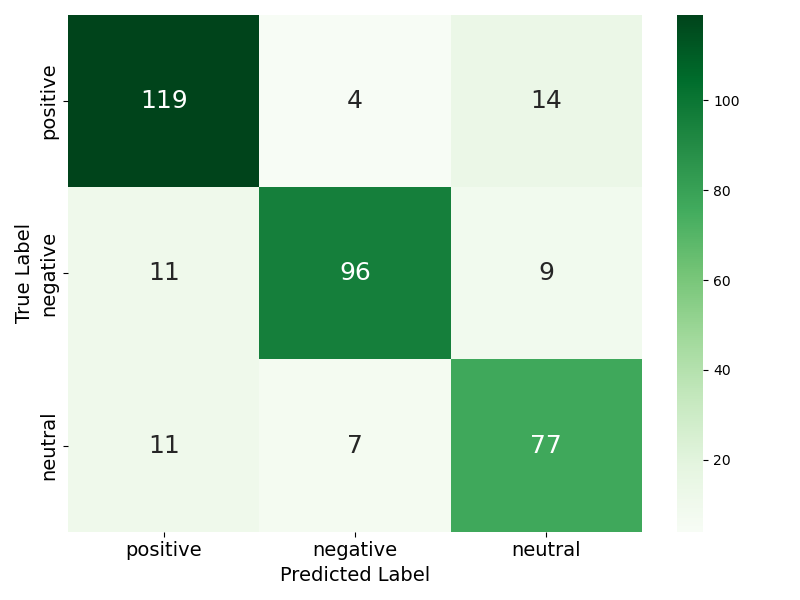}
}
\hfil
\subfloat[Restaurant Confusion Matrix\label{fig:restaurant_confusion}]{
    \includegraphics[width=0.3\textwidth]{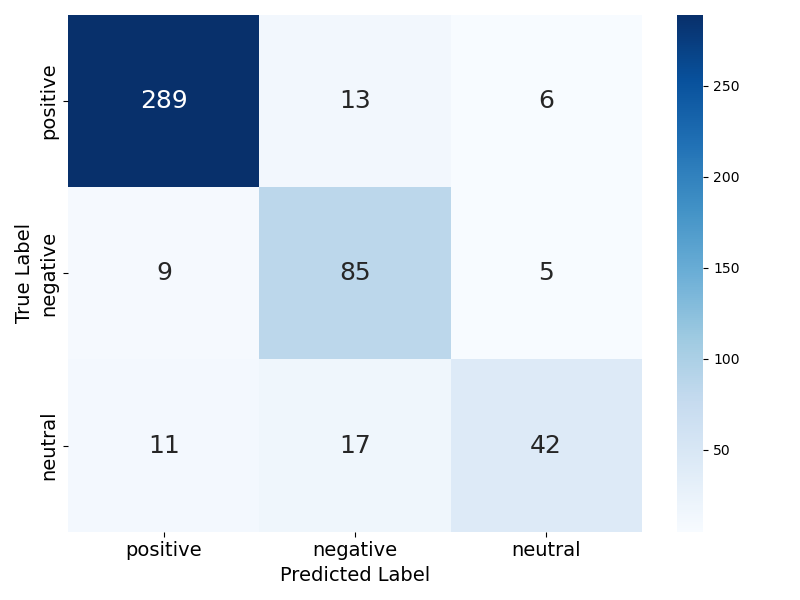}
}
\hfil
\subfloat[Twitter Confusion Matrix\label{fig:twitter_confusion}]{
    \includegraphics[width=0.3\textwidth]{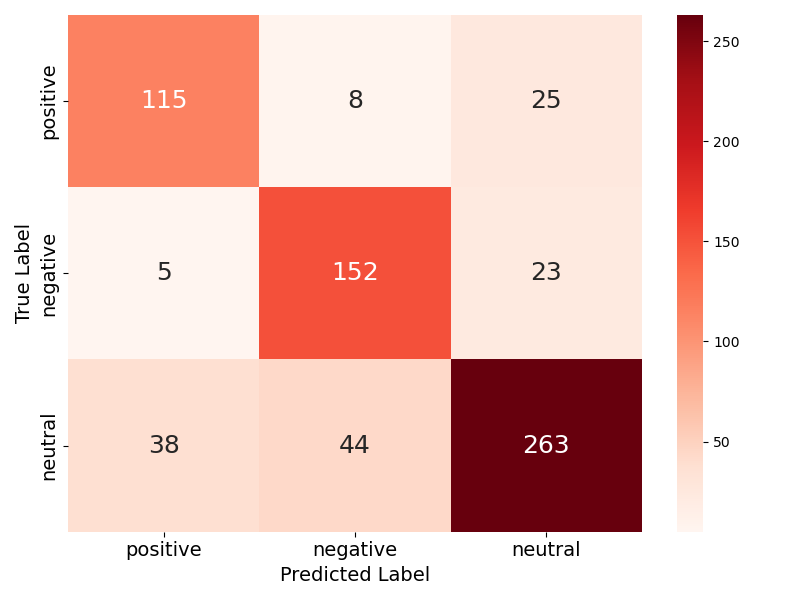}
}
\caption{Confusion Matrices for Different Datasets}
\label{fig7}
\end{figure*}

\paragraph{Versus structure-enhanced / fusion models.}
Compared with HPEP-GCN and YORO~\cite{b27,b22}, OTESGN attains \textbf{80.52} on \textsc{Laptop14} (\textbf{+1.42 pp} over HPEP-GCN).
Its learnable syntactic masks and gated fusion adaptively weight structural and semantic signals beyond template-driven schemes.
On \textsc{Twitter}, it achieves \textbf{78.17} (\textbf{+1.01 pp} over LWEDA-ACT~\cite{b36}).

\subsection{Ablation Study}

We perform one-component-at-a-time ablations and a joint ablation (GA+OT).
Table~\ref{tab3} presents Accuracy and Macro-F1 (\%) across the ablation configurations.
All deltas are reported in percentage points (pp) relative to the full OTESGN.
We denote SM (syntactic masking), GA (gated attention/fusion), OT (optimal transport), and CL (contrastive loss).

\paragraph{w/o SM.}
Removing SM leads to large drops on \textsc{Rest14} (Acc $-4.30$~pp, F1 $-4.47$~pp) and \textsc{Twitter} (Acc $-7.13$~pp, F1 $-8.97$~pp).
The numerically larger decrease on \textsc{Twitter}—a corpus with looser syntax—suggests SM helps capture latent structures in informal text.

\paragraph{w/o GA.}
We observe declines on \textsc{Rest14} (Acc $-3.27$~pp) and \textsc{Twitter} (Acc $-3.12$~pp), while \textsc{Laptop14} shows a small Acc drop ($-0.56$~pp) and a marginal F1 change ($+0.16$~pp).
This pattern indicates GA is generally beneficial for semantic modeling, whereas in attribute-oriented \textsc{Laptop14}, reducing the structural channel can occasionally bypass syntactic noise.

\paragraph{w/o OT.}
Removing OT yields the strongest cross-domain degradation (e.g., \textsc{Rest14} Acc $-6.61$~pp; \textsc{Twitter} F1 $-6.61$~pp),
indicating the OT module contributes substantially to aspect--opinion alignment and robustness to long-range or noisy dependencies.

\paragraph{w/o GA+OT.}
The joint removal shows (i) on \textsc{Rest14}, the drop ($-6.85$~pp Acc) is close to w/o OT ($-6.61$~pp), suggesting OT accounts for most of the gain;
(ii) on \textsc{Laptop14}, the combined drop ($-5.02$~pp Acc) exceeds either single removal, indicating a weak positive interaction;
(iii) on \textsc{Twitter}, the larger decrease ($-12.43$~pp Acc) implies both channels provide complementary robustness under informal text.

\paragraph{w/o CL.}
The F1 decrease is larger on \textsc{Twitter} ($-4.21$~pp) than on \textsc{Rest14}/\textsc{Laptop14} ($-1.89$/$-1.35$~pp),
consistent with CL mitigating noise from informal expressions.

\paragraph{Takeaways.}
Across datasets, removing any single component consistently degrades performance; OT exhibits the largest average impact.
The evolution of the learnable weight $\beta$ in Fig.~\ref{fig6} (converging to $\sim$0.10--0.30) is consistent with the ablation trends, suggesting a stable contribution of the OT branch.

\subsection{Case Study}
Comprehensive case analysis of misclassified samples (Table~\ref{tab4}) reveals the primary error pattern based on the confusion matrix Fig.~\ref{fig7}, the primary error pattern of the OTESGN model lies in misclassifying positive and negative sentiments as neutral. This is mainly attributed to the absence of explicit sentiment markers and the implicit nature of deep semantic associations in the input. As illustrated in Cases 1 and 2, the lack of directly expressed polarity words leads the model to adopt a more conservative, neutral prediction. Additionally, we observe instances of negative sentiments being misclassified as positive, such as in Case 3. This indicates that when subjective opinions are expressed through discontinuous or nested structures, the model struggles to capture the underlying sentiment reversal effectively. These findings highlight current limitations in implicit sentiment recognition and complex syntactic parsing. Future work could explore the integration of context-aware mechanisms or enhanced syntactic dependency modeling to address these challenges.

\subsection{Visualization Analysis}

\begin{figure}[t]  
\centering
\includegraphics[width=\columnwidth]{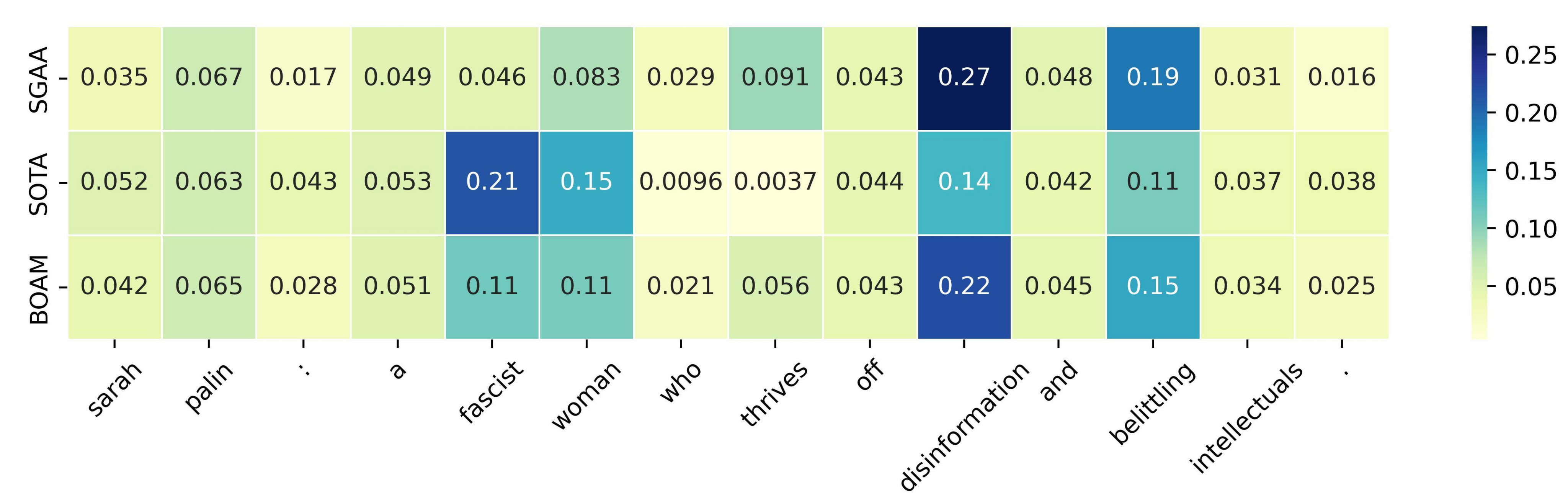} 
\caption{Word-Level Attention Heatmap}
\label{fig8}
\vspace{-\baselineskip} 
\end{figure}

As illustrated in the attention heatmap in Fig.~\ref{fig8}, the structure-aware graph attention effectively focuses on words syntactically close and semantically related to the aspect term \texttt{"Sarah Palin"} such as \texttt{"disinformation"} (weight=0.27) and \texttt{"belittling"} (weight=0.19), demonstrating strong structural guidance. 

\begin{figure}[t]  
\setlength{\belowcaptionskip}{-10pt} 
\centering
\includegraphics[width=\columnwidth]{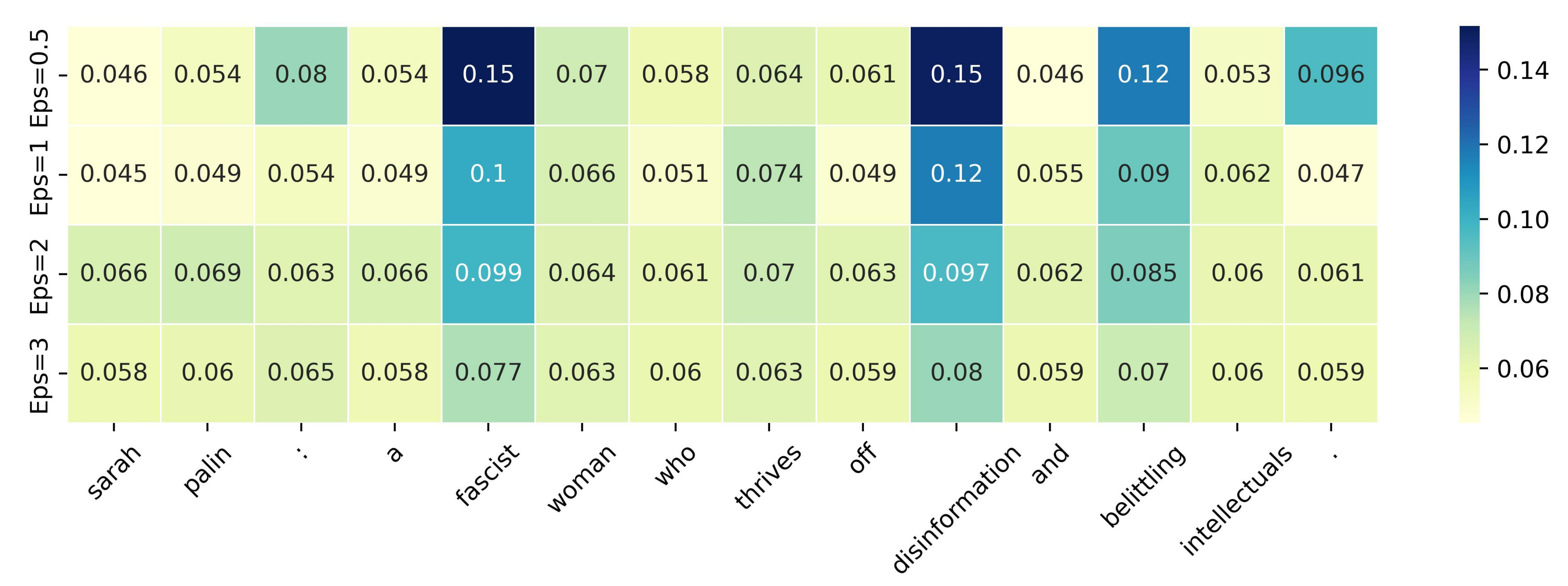} 
\caption{Effect of Hyperparameter Eps on OT Attention Scores}
\label{fig9}
\end{figure}

\begin{figure}[t]
\centering
\subfloat[Effect of Dropout Rate on Model Accuracy.]{
    \includegraphics[width=0.95\linewidth, height=5cm]{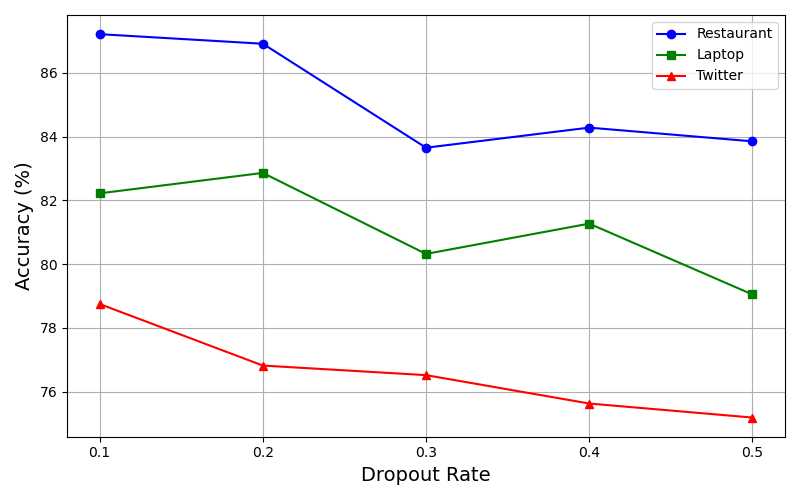}
}
\vspace{1em}

\subfloat[Effect of OT Attention Layer Depth on Model Accuracy.]{
    \includegraphics[width=0.95\linewidth, height=5cm]{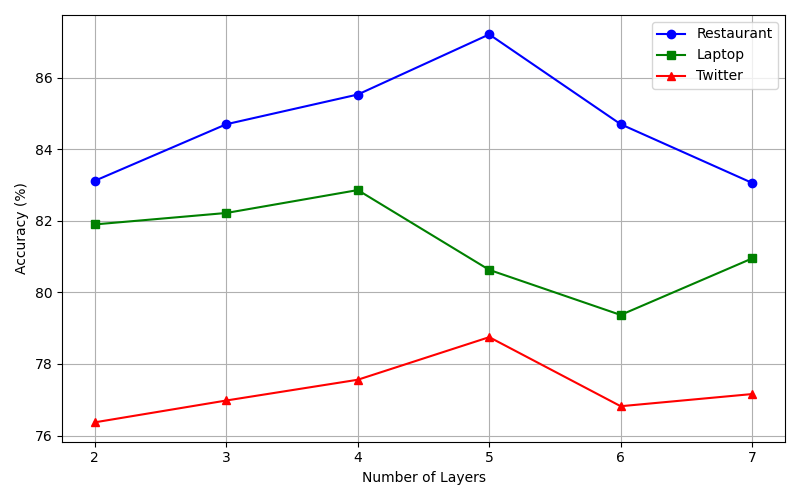}
}
\caption{Dropout Regularization and OT Attention Dynamics Across Network Layers}
\label{fig10}
\end{figure}

In contrast, the OT-based attention effectively complements structural limitations by assigning high weights to semantically aligned yet potentially distant words, such as \texttt{"fascist"} (weight=0.21) and \texttt{"disinformation"} (weight=0.14), highlighting its ability to capture fine-grained semantic alignment. The final aggregated attention integrates both strengths, assigning high importance to key sentiment words like \texttt{"fascist"} (0.11), \texttt{"disinformation"} (0.22), and \texttt{"belittling"} (0.15), leading to a more comprehensive focus on aspect-relevant expressions and thereby enhancing the interpretation of complex emotional nuances.

Fig.~\ref{fig9} depicts OT attention score distributions under different $\epsilon^k$ values. At smaller $\epsilon^k$, attention becomes sharper, strongly emphasizing key semantic words like \texttt{"fascist"} and \texttt{"disinformation"} (weights=0.15). However, this concentration also inadvertently amplifies weights for semantically minimal tokens (e.g., \texttt{":"}, \texttt{"."}). Conversely, increasing $\epsilon^k$ (e.g., $\epsilon^k$=3) smoothes the distribution, reducing variance and the model's discriminative power between common and key words. Therefore, careful $\epsilon^k$ selection is crucial to effectively guide OT attention towards essential semantics, balancing focused emphasis with noise resilience.

\subsection{Hyperparameter Sensitivity Analysis}
We conducted sensitivity analyses on two key hyperparameters of the OTESGN model—dropout rate and number of OT attention layers—across multiple domain datasets. As depicted in Fig.~\ref{fig10} (a), optimal accuracy for the Rest14 and Twitter datasets was achieved with a low dropout rate of 0.1, suggesting these domains require minimal regularization to preserve original features for capturing fragmented linguistic cues. Conversely, Laptop14 peaked at an 82.86\% accuracy with a 0.2 dropout rate, indicating a greater need for regularization to mitigate oversensitivity to specific expressions. Regarding OT attention layer count (Fig.~\ref{fig10} (b)), Rest14 accuracy showed an inverted V-shaped trend, peaking at approximately 87\% at layer 5 before declining, suggesting potential overfitting or reduced training efficiency beyond this depth. Laptop14 accuracy rose slowly to its peak at layer 4, declined steadily, but then recovered slightly from layer 6, implying that medium-depth networks best capture its longer-range dependencies, while excessive layers cause semantic dispersion. Twitter accuracy improved gradually, peaked at layer 5, then fell sharply before stabilizing; this indicates shallow semantic features dominate in social media text, with deeper networks potentially inducing noise memorization.

\section{Conclusion}
This paper presented OTESGN, an Optimal Transport-Enhanced Syntactic–Semantic Graph Network for aspect-based sentiment analysis. By integrating syntactic constraints with distributional alignment, the model captures both global dependencies and fine-grained semantic associations. Specifically, Syntactic Graph-Aware Attention encodes structural relations, while Semantic Optimal Transport Attention formulates aspect–opinion matching as a distribution problem solved via the Sinkhorn algorithm. An Adaptive Attention Fusion mechanism balances these signals, and contrastive regularization improves robustness. Experiments on Rest14, Laptop14, and Twitter show that OTESGN consistently outperforms or matches state-of-the-art baselines, with notable gains on Laptop14 and Twitter. Ablation and visualization further highlight the role of optimal transport in enhancing alignment and mitigating noise. 

\section{Limitations}
On Rest14, OTESGN is competitive but does not surpass the strongest baseline. Likely factors include a stronger reliance on explicit syntax under standardized reviews and occasional structural mismatches from the dual-channel attention. Additionally, we observe errors in cases involving implicit polarity, where subtle sentiment cues are not explicitly expressed in the text. Another limitation is the moderate overhead of the OT solver relative to dot-product attention; we mitigate this by tuning the entropy regularization coefficient $\epsilon^k$ to balance coupling sharpness and convergence. Future work will explore adaptive syntax extraction and the integration of event/knowledge priors to further improve robustness.

\end{document}